\documentclass[10pt,twocolumn,letterpaper]{article}

\usepackage{graphicx}
\usepackage{amsmath}
\usepackage{amssymb}
\usepackage{booktabs}
\usepackage[T1]{fontenc}
\usepackage{babel}
\usepackage{float}
\usepackage{mathtools}
\newcommand\numberthis{\addtocounter{equation}{1}\tag{\theequation}}
\usepackage[pagebackref,breaklinks,colorlinks]{hyperref}

\begin{document}
% \begin{frontmatter}
\title{Audio-visual video face hallucination with frequency supervision and cross modality support by speech based lip reading loss}

%Automatic speech recognition inspired lip reading loss for cross modal audio-visual video face hallucination network with frequency supervision
% \author[1]{Shailza \snm{Sharma}\corref{cor1}} 
% \cortext[cor1]{Corresponding author}
% \ead{ssharma_phd18@thapar.edu}
% \author[2]{Abhinav \snm{Dhall}}
% \author[1]{Vinay \snm{Kumar}}
% \author[3]{Vivek \snm{S\. Bawa}}
% \address[1]{Thapar Institute of Engineering and Technology, India}
% \address[2]{Indian Institute of Technology, Ropar}
% \address[3]{Oxford Brookes University, UK}
\author{Shailza Sharma\\
Thapar Institute of Engineering and Technology\\
Punjab, India\\
{\tt\small ssharma\_phd18@thapar.edu}
\and
Abhinav Dhall\\
Indian Institute of Technology Ropar\\
India\\
{\tt\small abhinav@iitrpr.ac.in}
\and
Vinay Kumar\\
Thapar Institute of Engineering and Technology\\
Punjab, India\\
{\tt\small vinay.kumar@thapar.edu}
\and
Vivek Singh Bawa\\
Oxford Brookes University\\
UK\\
{\tt\small vsingh@brookes.ac.uk}
}
\maketitle

%%%%%%%%% ABSTRACT
\begin{abstract}
Recently, there has been numerous breakthroughs in face hallucination tasks. However, the task remains rather challenging in videos in comparison to the images due to inherent consistency issues. The presence of extra temporal dimension in video face hallucination makes it non-trivial to learn the facial motion through out the sequence. In order to learn these fine spatio-temporal motion details, we propose a novel cross-modal audio-visual Video Face Hallucination Generative Adversarial Network (VFH-GAN). The architecture exploits the semantic correlation of between the movement of the facial structure and the associated speech signal. Another major issue in present video based approaches is the presence of blurriness around the key facial regions such as mouth and lips - where spatial displacement is much higher in comparison to other areas. The proposed approach explicitly defines a lip reading loss to learn the fine grain motion in these facial areas. During training, GANs have potential to fit frequencies from low to high, which leads to miss the hard to synthesize frequencies. Therefore, to add salient frequency features to the network we add a frequency based loss function. The visual and the quantitative comparison with state-of-the-art shows a significant improvement in performance and efficacy.
\end{abstract}
% \end{frontmatter}
%%%%%%%%% BODY TEXT
\section{Introduction}
\label{sec:intro}
Face super resolution (SR) (also known as face hallucination) refers to the problem of
generating high resolution face images from its low resolution counterpart. Resolution enhancement of the facial images has a wide range of applications in different fields of computer vision, such as face recognition \cite{zeng2021survey},
deep fake generation and detection \cite{yadav2019deepfake,remya2021detection},
emotion detection \cite{cai2021identity}, face identification \cite{lin2021fpgan}
and face alignment \cite{wan2020robust}, etc.

There are numerous deep learning algorithms that effectively solves the face hallucination problem in images \cite{hu2020face, he2022gcfsr, fan2020facial}. Despite many breakthroughs in facial super resolution, there are numerous challenges in Video Face Hallucination (VFH). Temporal and spatial information are the prerequisites for VFH models. Since, single image is considered in face hallucination problems - only spatial information has been predominantly used in most of the state-of-the-art (SOTA) approaches. On the other hand, to solve the VFH problem, temporal motions of the spatial features is a key component. Various techniques have been proposed to incorporate the temporal information in VFH models, such as, 3D convolution networks used by \cite{qiu2017learning} to add temporal consistency to the network for VFH. But these methods require very high computational resources. Another methods \cite{caballero2017real,jo2018deep} include stacking multiple low resolution frames from the videos and pass them to the deep neural network model. Since, these methods  naively fuse multiple frames at the input level, no implicit leaning mechanism is used to learn the temporal relationship between the spatial features. Semantic priors, facial parsing maps
and 3D priors also assists the VFH networks to generate superior results. Yet, the requirement of pre-trained networks to obtain the prior information leads to extra computation. %Please rewritw, unable to understand.

One major limitation of generative models is their ability to fit a range of
frequencies in a pattern during the training \cite{xu2019frequency}. The models prioritise a smaller band of frequencies instead of learning the whole spectrum. Hence, the salient local features of the regions are harder to learn as they are generally absent in the low resolution image. This issue has been overlooked in the GAN based face super resolution literature. Most of the SOTA use objective function that only optimize the spatial domain features. Consequently, learned feature representations do not contain the high frequency detail leading to lower sharpness of the visual output. Furthermore, current VFH approaches have blurriness in the regions with higher motion, such as mouth, lips, teeth, etc. The root cause of the problem is absence on implicit mechanism to learn temporal dependence between the frame. The higher structural complexity of the region is also partially responsible for the degradation of the quality.

The audio and video have very high semantic correlation. This correlation can be used as a implicit semantic supervision in the VFH networks. Speech signal based objective can help in learning the temporal consistency between the frames. Moreover, a lot of identity, age and gender information are also carried in audio signals \cite{oh2019speech2face}.
Hence, in the very low resolution images where gender and identity information are hard to retrieve - audio signal can play a critical role in generating such details.

In this paper, we exploit the correlation between the frequency spectrum of the speech signal and the motion of spatial regions (like, mouth and lips). We developed a multi-modal GAN architecture that uses speech and video modalities to further enhance the quality of super resolution (SR) image. The proposed method uses two feature encoder backbones in the generator network. First backbone extract the spatial features from the facial image while second extracts the audio features at corresponding time step. An axial self attention mechanism is used for the feature fusion based feature fusion between both the modalities. 
We also use a frequency based loss function (calculated using 2D discrete Fourier transform) in order to generate the high frequency details which are otherwise difficult to generate. We also present the lip reading loss to resolve the issue of blurriness around the mouth and lips region. Instead of directly minimizing the distance between generated image and the ground-truth image - distance between their corresponding feature maps (extracted from the intermediate layers of the pre-trained lip-reading network) are minimized. The proposed loss function propels the video
face hallucination network (VFHN) to generate facial images with very fine mouth region.
Main contributions are as follows:
\begin{itemize}
\item A cross-modal learning mechanism for audio-visual data is explored to learn the fine spatial-temporal motion details and facial identity information.
\item An explicitly defined lip reading loss is used to learn a fine grain motion in the frames to remove the blurriness in key facial regions.  
\item A Fourier transform based frequency domain loss is also applied to add the salient frequency features.
\item Visual results, quantitative numbers along with their edge restoration
number (metric used to examine the restoration quality of edges in
videos) shows superiority of proposed work over other SOTA
methods. 
\end{itemize}

\section{Related work}

$\mathbf{Face}$ $\mathbf{super}$ $\mathbf{resolution}$ $\mathbf{using}$
$\mathbf{GANs}$: Earlier works in the field of FSR using GANs include
numerous renowned works. Ultra resolving facial images with the help
of discriminator network is one of them \cite{yu2016ultra}. In this
model, discriminator is fed with the image generated from the generator
network along with the ground truth high resolution image, compelling
the generator to mimic high resolution images. To mitigate the training
difficulties in GANs and stabilize the training process, Chen et al.
\cite{chen2017face} proposed a Wasserstein distance as a training
metric for FSR system.

Ko et al. \cite{ko2021multi} argued that recent GAN based methods
require extra information along with the low resolution image to generate
images with fine perceptual details. But they utilized only a low
resolution image with its edge information at various scales to generate
high resolution image. Most GAN based methods use bicubic kernels
to obtain low resolution image from the high resolution image for
training. So, the training dataset does not follow the natural degradation
process which affects the performance of GAN based methods on realistic
low resolution images. To address this problem Aakerberg et al. \cite{aakerberg2022real}
introduced different types of noises in low resolution images for
training dataset. Xiaobin et al. \cite{hu2020face} used 3D priors that carry facial expression, pose and identity information to obtain sharp facial features. An auxiliary depth estimation branch is proposed by Fan et al. \cite{fan2020facial} in aggregation with the super resolution branch to add the depth information in facial image. He et al. \cite{he2022gcfsr} proposed encoder-generator based multi module architecture to mitigate the requirement of facial priors in face super resolution. By utilizing basic functions like bidirectional
backpropagation and feature alignment in a systematic way. Although these prior information based methods have achieved great success in VFH, yet the requirement of pre-trained models increases the computational complexity of the overall method to an enormous rate. Chan et
al. \cite{chan2021basicvsr} achieved PSNR higher than other competing
networks. Yet higher PSNR does not necessarily indicates more pleasing
results visually \cite{ledig2017photo}.

$\mathbf{Audio}$-$\mathbf{visual}$ $\mathbf{based}$ $\mathbf{learning}$
$\mathbf{methods}$: Semantic correlation between the audio and visual
information is utilized in numerous computer vision problems \cite{belin2004thinking,hu2019deep}.
For example, Tian et al. \cite{tian2020unified} used the aural information
for the comprehensive study of scene. \cite{xuan2020cross} and \cite{wu2019dual}
proposed an attention mechanism between the cross-modal aural-visual
network to abolish the temporal inconsistency during localization
of events and analyze the longer videos with prominent information,
respectively. By exploring the relation between speech and visual
representation, Wen et al. \cite{wen2019face} and Oh et al. \cite{oh2019speech2face}
proposed GAN based networks that generates facial images from the
speech using physical attributes like identity matching, age and gender
etc. Chen et al. \cite{chen2019hierarchical} converted the aural
signals into a complex structure which corresponds to facial landmarks
and using these landmarks facial images are generated.

Zhang et al. \cite{zhang2020davd} embedded the aural information
in the CNN architecture to enhance the facial videos and remove compression
deformities. Along with low resolution image encoder, the audio encoder
is used by \cite{meishvili2020learning} for image super resolurion.
Feature maps obtained by amalgamating the hierarchical features of
both the encoders are applied to the decoder, resulting in a high
resolution image. Taking inspiration from cross-modal architectures in video super resolution, we propose an audio-visual module for face hallucination in videos that maps the spatial displacement across the video frames and enforces the model to learn identity and gender information in the generated images.

\section{Proposed Methodology}

Often, facial videos have audios associated with them. Yet, the semantic
correlation between the audio features and facial frames remained
unexplored in the previous literature of VFH. Therefore, a cross-modal
architecture to understand the significance of aural features in facial
videos is proposed. In addition, to explicitly
add salient frequency features, a Fourier transform based frequency
domain loss is proposed. To address the issue of blurriness of mouth region and ensure the temporal consistency across the video frames, a lip
reading based loss function in introduced. The loss compels our video
face hallucination network to generate the final image with fine texture
around the mouth region and consistency across the frames. In this
section, the proposed generator architecture and loss functions to
optimize the proposed model are explained in detail. 

\subsection{Overview}

\begin{figure*}[!ht]
  \centering
   %\includegraphics[width=0.8\linewidth]{egfigure.eps}
   \includegraphics[scale = 0.6]{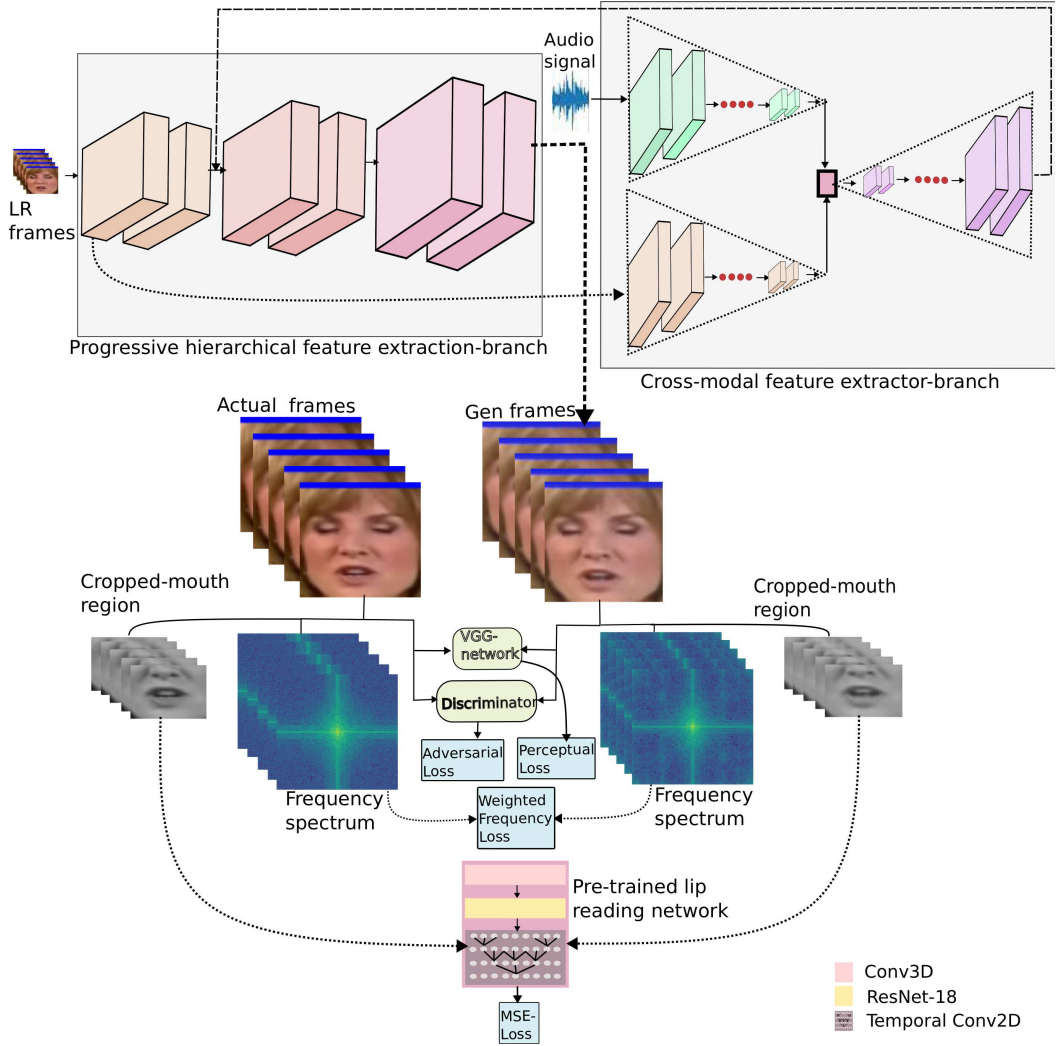}
   \caption{VFH-GAN: final output is generated by progressive hierarchical feature extraction branch PHFE-B). Audio features are extracted from the cross-modal feature extractor branch (CMFE-B)  and embedded into PHFE-B}
   \label{fig:Generator-architecture}
\end{figure*}

% \begin{figure}[ht]
% \centering

% \includegraphics[width = \columnwidth]{arch_m.eps}
% \caption{Generator architecture. \label{fig:Generator-architecture}}
% \end{figure}

The aim to perform video face hallucination is to generate a high
resolution video frame ($\mathit{f}_{t}^{hr}$) from corresponding
low resolution video frame ($\mathit{f}_{t}^{lr}$), its neighboring
low resolution frames ($\mathit{f}_{t-j}^{lr}$,... $\mathit{f}_{t-1}^{lr},$$\mathit{f}_{t+1}^{lr}$,..$\mathit{f}_{t+j}^{lr}$)
and audio features ($a_{f}$).  Here, $t$ refers to the time step
at which face hallucination is performed and $2j$ are the neighboring
LR frames helping the chosen LR frame to generate a high resolution
video frame. 

CNNs perform better with image data, therefore,
rather than using audio features directly in spatial domain, we transformed
them into their frequency representation (spectrum) using short
time Fourier transform ($a_{t}=STFT(a_{f})$ ). This spectrum is further
converted into mel-spectograms ($a_{m_{-}s}$) for better perception
and used as an input to our VFHN (refer eq \ref{eq:}). 
\begin{equation}
  \mathit{f}_{t}^{sr}=\zeta((\mathit{f}_{t}^{lr},\{\mathit{f}_{i}^{lr}\}_{i=t-j}^{t+j}),a_{m_{-}s};\emptyset_{w_{v},b_{v}},\emptyset_{w_{a},b_{a}})
  \label{eq:}
\end{equation}

% \begin{equation}
% \mathit{f}_{t}^{sr}=\zeta((\mathit{f}_{t}^{lr},\{\mathit{f}_{i}^{lr}\}_{i=t-j}^{t+j}),a_{m_{-}s};\emptyset_{w_{v},b_{v}},\emptyset_{w_{a},b_{a}})\label{eq:}
% \end{equation}

here, $\mathit{f}_{t}^{sr}$ is the high resolution generated video
frame and $\zeta$ represents the VFHN. $\emptyset_{w_{v},b_{v}}$
and $\emptyset_{w_{a},b_{a}}$are the weights and bias parameters
of video based module and audio based module, respectively. Both these
network parameters are optimized by reducing the loss value calculated
between the ground-truth high resolution image and the generated image
from VFHN. 

Proposed architecture is depicted in Figure \ref{fig:Generator-architecture}.
The motivation for progressive hierarchical feature extractor branch
(PHFE-B) is taken from \cite{sharma2022frequency}. This branch comprises
of two modules. The first modules rely on the complex filter formation,
where input feature maps are applied to different convolution layers
with increasing filter sizes ($1\times1,3\times3,5\times5,7\times7$).
Different size kernels allow the network to assimilate primary or
the local attributes as well as hierarchical features. Second module
is channel attention module, where depthwise convolution layers are
utilized, establishing a computationally efficient architecture. Additionally,
squeeze and excitation block is used in this module, enforces the
network to emphasize on extracting feature maps from channels which
are more affluent then the others. 

As shown in Figure \ref{fig:Generator-architecture}, cross-modal
feature extraction branch splits into two branches. First branch takes
feature maps from the primary convolution layers of PHFE-B as input
($\xi(f_{t}^{lr'};\emptyset_{w_{v'},b_{v'}}))$ and audio signal is
applied as an input for the second branch ($\psi(a_{m_{-}s};\emptyset_{w_{a},b_{a}})$).
Feature embeddings obtained from these two branches are concatenated
and converted to another latent vector ($f^{lv'}$) (refer eq. \ref{eq:-9}).
This latent vector is applied to axial attention layer to sustain
the long range dependencies and give importance to feature vectors
with more relevance then the others and resulting in $f^{a}$ (refer
eq. \ref{eq:-10}).

\begin{equation}
    f^{lv'}=\varPsi\{(\xi(f_{t}^{lr'};\emptyset_{w_{v'},b_{v'}})+\psi(a_{m_{-}s};\emptyset_{w_{a},b_{a}}));\emptyset_{w_{lv},b_{lv}}\}
  \label{eq:-9}
\end{equation}

\begin{equation}
    f^{a}=axial_{-}atten(f^{lv'})
    \label{eq:-10}
\end{equation}

Obtained feature vector ($f^{a}$) is added with the feature maps
from PHFE-B resulting in high resolution facial image generated with
correct identity and gender information.

\subsection{Loss function}

Final video frame ($\mathit{f}_{t}^{sr}$) is generated by optimizing
the overall loss function ($L^{lf}$), calculated between the ground-truth
video frame ($\mathit{f}_{t}^{gt}$) and generated video frame ($\mathit{f}_{t}^{sr}$)
over $N$ training data samples. This loss function is composed of
few weighted loss functions explained in detail in this section.

\subsubsection{Lip reading loss function}

Accurately identifying the lip movements is elemental for visual speech
recognition. Hence, this lip-reading application involves explicating
the movement of teeth, tongue and lips. Another requisite for correct
lip-reading is temporal dependency across the video frames. These
attributes of lip-reading networks can assist our video face hallucination
network to generate frames with very fine texture of mouth region
and maintain temporal consistency across video frames. We used pre-trained
lip-reading network \cite{martinez2020lipreading} to extract these
texture rich feature embedding. The overview of the lip-reading model
we used is as follows. 

Firstly, a sequence of video frames with cropped mouth region are
applied as an input to the 3D convolution layer (refer eq. \ref{eq:-1}). 

\begin{equation}
   f^{sr'}=Conv3D(\{f_{i}^{sr}\}_{i=t-j}^{t+j})
   \label{eq:-1}
\end{equation}

here, $\{f_{i}^{sr}\}_{t-j}^{t+j}\epsilon$$\mathbb{R}^{b*k*h*w}$,
$b$ is the batch size, $k$ represents the number of frames in the
video sequence, height and width of video frame is denoted by $h$
and $w$, respectively. $f^{sr'}$ is the feature embedding obtained
after applying 3D convolution layer to the video sequence. 

The output features obtained are applied to ResNet-18 ($f^{sr''}=ResNet(f^{sr'})$)
by reshaping the feature maps from $\boldsymbol{b*k*h*w}$ to $\boldsymbol{B*h*w}$.
Here, $B$ refers to the number of frames concatenated over batch dimension and
$f^{sr''}$ represents the visual feature maps obtained from ResNet-18
model. 

Obtained visual feature maps are applied to multi-scale temporal convolution
network, which uses dilated convolution layers in place of convolution
layer. Multi-scaling assists the network to map short term as well
as long term dependencies. 

We used this lip-reading network to extract high level feature embedding.
Firstly, we cropped the mouth region from $\{f_{i}^{sr}\}_{t-j}^{t+j}$
and $\{f_{i}^{gt}\}_{t-j}^{t+j}$ , converted them into gray scale
and then pass these video sequences from pre-trained lip-reading network
(refer eq \ref{eq:-2}). 

% \begin{center}
% \begin{equation}
% L_{vgg/i.j}^{f}=\frac{1}{W_{i,j}H_{i,j}}\stackrel[a=1]{W_{i,j}}{\sum}\stackrel[b=1]{H_{i,j}}{\sum}(\varphi_{i,j}(i^{hr})_{a,b}-\varphi_{i,j}(G_{\xi_{g}}(i^{lr}))_{a,b})^{2}\label{eq:loss2}
% \end{equation}
% \par\end{center}

\begin{align*}
L_{L_{-}Rd/l.m}^{lf_{1}}=\frac{1}{w_{l,m}h_{l,m}}\sum_{\substack{x={}1}}^{w_{l,m}}\sum_{\substack{y={}1}}^{h_{l,m}}
\aleph_{l,m}(\{f_{i}^{gt}\}_{i=t-j}^{t+j})_{x,y}-\\ \aleph_{l,m}(\{f_{i}^{sr}\}_{i=t-j}^{t+j})_{x,y} \numberthis
\label{eq:-2}
\end{align*}

% \begin{align*}
% L_{L_{-}Rd/l.m}^{lf_{1}}=\frac{1}{w_{l,m}h_{l,m}}\stackrel [x=1]{w_{l,m}}{\sum}\stackrel[y=1]{h_{l,m}}\\
% {\sum}\aleph_{l,m}(\{f_{i}^{gt}\}_{i=t-j}^{t+j})_{x,y}-\aleph_{l,m}(\{f_{i}^{sr}\}_{i=t-j}^{t+j})_{x,y} \numberthis
% \label{eq:-2}
% \end{align*}

here, $L_{L_{-}Rd/l.m}^{lf_{1}}$ is the lip-reading loss. $\aleph_{l,m}(\{f_{i}^{gt}\}_{i=t-j}^{t+j})$
and $\aleph_{l,m}(\{f_{i}^{sr}\}_{i=t-j}^{t+j})$ are the feature
embedding obtained from the intermediate layers of lip-reading network.
$L_{1}$loss is calculated between these feature embedding. This loss
enforces our VFHN to generate images with high texture in mouth region
and maintain temporal consistency across the sequence of videos.

\subsubsection{Weighted Fourier Frequency Loss}

Frequency representation of images give better perception of artifacts
present in an image \cite{jiang2021focal}. Missing high frequency
information in the frequency domain will lead to ringing artifacts
in the image in spatial domain. Whereas, if solely, the high frequency
information is present in the image, it corresponds to the boundaries and
edges in the visual domain. Checkerboard artifacts can be seen in
the spatial domain with the usage of band stop filtering in frequency
representation. From the above information, we can conclude that missing
frequencies in the frequency representation of an images are equivalent
to various artifacts in the spatial domain. Hence, incorporating these
missing frequencies in the frequency domain will lead to better perceptual
quality in spatial domain. 

We are using 2D discrete fourier transform (DFT) for the frequency
representation of an image (refer eq. \ref{eq:-3}, \ref{eq:-4}).

\begin{equation}
  \begin{aligned}
  \digamma(u,v)=\sum_{\substack{x={}0}}^{w-1}\sum_{\substack{y={}0}}^{h-1}(P_{i}(x,y))_{i=t-j}^{t+j}. (cos2\pi  (\frac{ux}{w}+\frac{vy}{h})- \\ isin2\pi(\frac{ux}{w}+\frac{vy}{h}))
  \label{eq:-3}
  \end{aligned}
\end{equation}
 
\begin{equation}
  \begin{aligned}
  \digamma^{\star}(u,v)=\sum_{\substack{x={}0}}^{w-1}\sum_{\substack{y={}0}}^{h-1}(P_{i}^{\star}(x,y))_{i=t-j}^{t+j}.(cos2\pi (\frac{ux}{w}+\frac{vy}{h})- \\ isin2\pi(\frac{ux}{w}+\frac{vy}{h}))
  \label{eq:-4}
  \end{aligned}
\end{equation}
 
 here, $P(x,y)$, $P_{i}^{\star}(x,y)$ are the pixel values of ground-truth
and generated video frames at $x$ and $y$ coordinates, respectively.
Frequency spectrum coordinates are represented by $u$ and $v$ and
its value by $\digamma^{\star}(u,v)$. Mean square error is calculated
between the ground-truth and generated image in the frequency domain
as shown in eq. \ref{eq:-5}.

\begin{equation}
L_{freq}^{l2}=\frac{1}{wh}\sum_{\substack{u={}0}}^{w-1}\sum_{\substack{v={}0}}^{h-1}|\digamma(u,v)-\digamma^{\star}(u,v)|^{2}\label{eq:-5}
\end{equation}

% \begin{equation}
% L_{freq}^{l2}=\frac{1}{wh}\stackrel[u=0]{w-1}{\sum}\stackrel[v=0]{h-1}{\sum}|\digamma(u,v)-\digamma^{\star}(u,v)|^{2}\label{eq:-5}
% \end{equation}

Generative models are more inclined towards generating easy frequencies
as compare to hard frequencies. Since, each frequency value have same
weightage and inherent biasing allows generative models to learn easy
frequencies better than the hard frequencies. During training, to
put more weightage to hard frequencies, a weight matrix (refer eq
. \ref{eq:-6}), similar to the shape of spectrum, is introduced which
adds non-uniformity to each frequency component in the cost function. 

\begin{equation}
m(u,v)=|\digamma(u,v)-\digamma^{\star}(u,v)|\label{eq:-6}
\end{equation}

$m(u,v)$ is the weight matrix having range $[0,1]$, where weights
near $0$ signifies the frequencies with more weightage and $1$ signifies
the frequency which is getting vanished. Therefore, frequencies which
are learned by the model easily are down-weighted. And hence, the
final weighted frequency cost function $(L_{freq}^{l2^{\ast}})$ is
shown in eq. \ref{eq:-7}.

\begin{equation}
L_{freq}^{lf_{2}^{\ast}}=\frac{1}{wh}\sum_{\substack{u={}0}}^{w-1}\sum_{\substack{v={}0}}^{h-1}{\sum}m(u,v)||\digamma(u,v)-\digamma^{\star}(u,v)||\label{eq:-7}
\end{equation}

\subsubsection{Final loss function}

Overall loss function is the weighted sum of lip-reading based loss
($L_{L_{-}Rd/l.m}^{lf_{1}}$), weighted frequency loss ($L_{freq}^{lf_{2}^{\ast}}$),
pre-trained vgg network based perceptual loss ($L_{vgg}^{lf_{3}}$)
\cite{ledig2017photo}, adversarial loss ($L_{adv}^{lf_{4}})$ (refer
eq. \ref{eq:-8}). 

\begin{equation}
L^{lf}=L_{vgg}^{lf_{3}}+\alpha L_{L_{-}Rd/l.m}^{lf_{1}}+\beta L_{freq}^{lf_{2}^{\ast}}+\gamma L_{adv}^{lf_{4}}\label{eq:-8}
\end{equation}

\section{Experiments}

\subsection{Datasets and Metrics}

Datasets with videos and corresponding audios for video face hallucination
are not publicly available. For training purpose, we selected LRW
dataset \cite{yang2019lrw}, a collection of videos having 500 words
spoken in various different sentences from more than $1000$ speakers.
We randomly selected videos from the 80 words (selected out of 500
words) and divided them in training, validation and testing. Facial
landmarks estimated using OpenFace \cite{amos2016openface} are used to obtain a square crop
and eliminate the undesired background from the frames. Cropped region
is resized into $128\times128$ and downsampled into $32\times32$
for low resolution input image. For testing purpose, other than LRW
dataset \cite{yang2019lrw}, we used Grid speech corpus \cite{alghamdi2018corpus}
and LRS2 \cite{afouras2018deep} datasets. All these datasets are
audio-visual datasets, as required by proposed network. 

In addition to the PSNR and SSIM (both calculated on Y channel after
transforming RGB image into YCbCr), we evaluated our results using
edge quality assessment metric (ERQA). This metric is used to estimate
the networks capability to restore the real information present in
the videos. 

\subsection{Ablation study}

\begin{figure}
\centering

\includegraphics[scale=0.5]{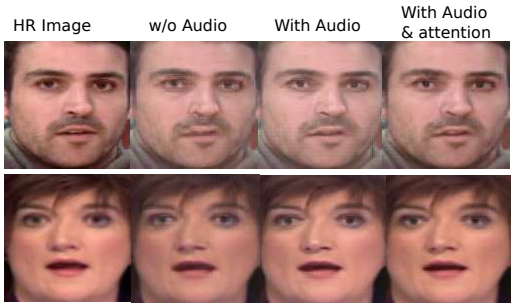}

\caption{Qualitative results evaluated on LRW test dataset showing the importance
of audio signals in the proposed network.\label{fig:Qualitative-results-evaluated}}
\end{figure}

$\mathbf{Audio}$ $\mathbf{signal}$ $\mathbf{analysis}$: In this
section, we studied the importance of audio waves in the proposed
architecture. At first, we use the PHFE-B, without the CMFE-B. From
the Figure \ref{fig:Qualitative-results-evaluated} and table \ref{tab:Quantitative-results-evaluated},
it is clear that there are little artifacts present in the facial keypoints of generated image and quantitative numbers are also less. After adding the CMFE-B, our results significantly improved, quantitative numbers - average PSNR ($\uparrow$$1.65$dB), SSIM ($\uparrow$$0.037$)  and ERQA ($\uparrow$$0.08$)) have increased.  To further
enhance the generated images, we added axial attention layer after
the CMFE-B, compelling the our network to emphasize on more important
feature vectors and improves the previous results.

\begin{table}
\centering

\begin{tabular}{c|c|c|c}
\toprule 
{\small{}Metric} & {\small{}w/o audio} & {\small{}audio} & {\small{}audio and attention}\tabularnewline
\midrule
{\small{}PSNR(db)} & {\small{}27.128} & {\small{}28.784} & {\small{}28.960}\tabularnewline

{\small{}SSIM} & {\small{}0.862} & {\small{}0.899} & {\small{}0.915}\tabularnewline
 
{\small{}ERQA} & {\small{}0.512} & {\small{}0.592} & {\small{}0.601}\tabularnewline
\bottomrule 
\end{tabular}

\caption{Quantitative results evaluated on LRW test dataset showing the importance
of audio signal in the proposed network.\label{tab:Quantitative-results-evaluated}}
\end{table}

\begin{figure}
\centering

\includegraphics[width = \columnwidth]{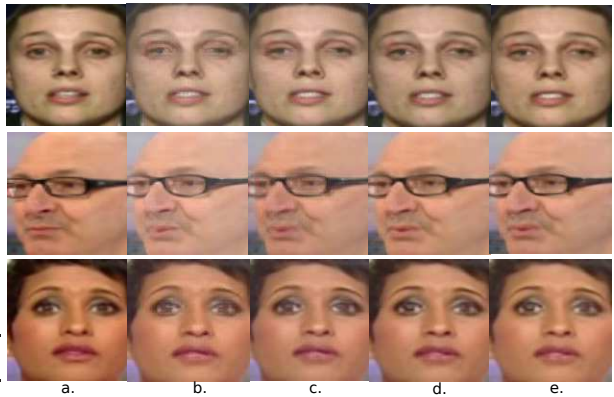}

\caption{Qualitative results for different loss functions: a. HR Image, b.
VFH-GAN with perceptual and adversarial loss, c. VFH-GAN with perceptual,adversarial
and weighted frequency loss, d. VFH-GAN with perceptual, adversarial
loss and lip-reading loss, e. VFH-GAN with perceptual, adversarial
loss, weighted frequency loss and lip-reading loss.\label{fig:Qualitative-results-for}}
\end{figure}

$\mathbf{Loss}$ $\mathbf{component}$ $\mathbf{analysis}$: We used
the final loss function as the combination of multiple loss functions
(refer \ref{eq:-8}). Initially, we applied the combination of perceptual
and adversarial loss (refer Figure \ref{fig:Qualitative-results-for}b
and table \ref{tab:Average-numbers-calculated}a. The results obtained
from the network have little color or texture problem. Further we
added weighted frequency element to loss function which corrects the
texture problem of the generated image (Figure \ref{fig:Qualitative-results-for}c
and table \ref{tab:Average-numbers-calculated}b). Still the generated
image has little artifacts. We added lip-reading loss with perceptual
and adversarial loss supervising our network to generate images with
high resolution around the mouth region and temporal consistency across
frames (Figure \ref{fig:Qualitative-results-for}d and table \ref{tab:Average-numbers-calculated}c)
. Therefore, the final image is generated by combining all these losses
resulting in Figure \ref{fig:Qualitative-results-for}e and table
\ref{tab:Average-numbers-calculated}d. There is significant improvement
in terms of texture and finer details in the final image. 

\begin{table}
\centering

{\small{}}%
\begin{tabular}{c|c|c|c|c}
\toprule 
{\small{}Metric} & {\small{}a.} & {\small{}b.} & {\small{}c.} & {\small{}d.}\tabularnewline
\midrule 
{\small{}PSNR} & {\small{}26.003} & {\small{}27.892} & {\small{}28.420} & {\small{}28.603}\tabularnewline

{\small{}SSIM} & {\small{}0.864} & {\small{}0.893} & {\small{}0.902} & {\small{}0.911}\tabularnewline
 
{\small{}ERQA} & {\small{}0.413} & {\small{}0.471} & {\small{}0.472} & {\small{}0.497}\tabularnewline
\bottomrule 
\end{tabular}{\small\par}

\caption{Average numbers calculated on LRS2 test dataset by using different
combinations of loss functions.\label{tab:Average-numbers-calculated}}
\vspace{-4mm}
\end{table}

\begin{figure}
\centering

\includegraphics[width = 0.35\columnwidth]{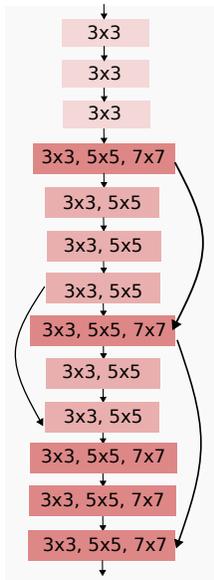}

\caption{mixnet block\label{fig:mixnet}}

\end{figure}

\begin{figure}
\centering

\includegraphics[width = 0.650\columnwidth]{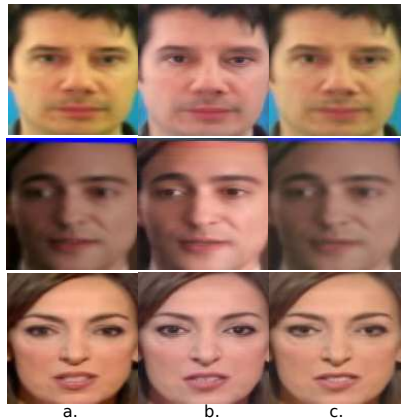}

\caption{Effect of different backbone architectures: a. High resolution image,
b. image generated with dice-net architecture as backbone architecture
and c. image generated with PVFH-B as backbone architecture.\label{fig:Effect-of-different}}

\end{figure}

$\mathbf{Different}$ $\mathbf{backbone}$ $\mathbf{architectures}$:
We also studied the effect of different backbone architectures. Firstly,
we used architecture (refer Figure \ref{fig:mixnet}) for the backbone of proposed
model. We used two mixnet blocks in each stage. Results obtained using
this architecture have high resolution with sharp features (refer
Figure \ref{fig:Effect-of-different}b). Yet, there is texture issue
in the generated images, color in the generated image does not match with
the ground-truth image. Then, we used the PFH-B \cite{sharma2022frequency}.
This architecture solves the texture problem and gives more accurate
results then the previous architecture. Therefore, for the final model
we used the PFH-B as backbone architecture.

\subsection{Comparison with the SOTA methods}

\begin{figure*}
\centering

\includegraphics[scale=0.45]{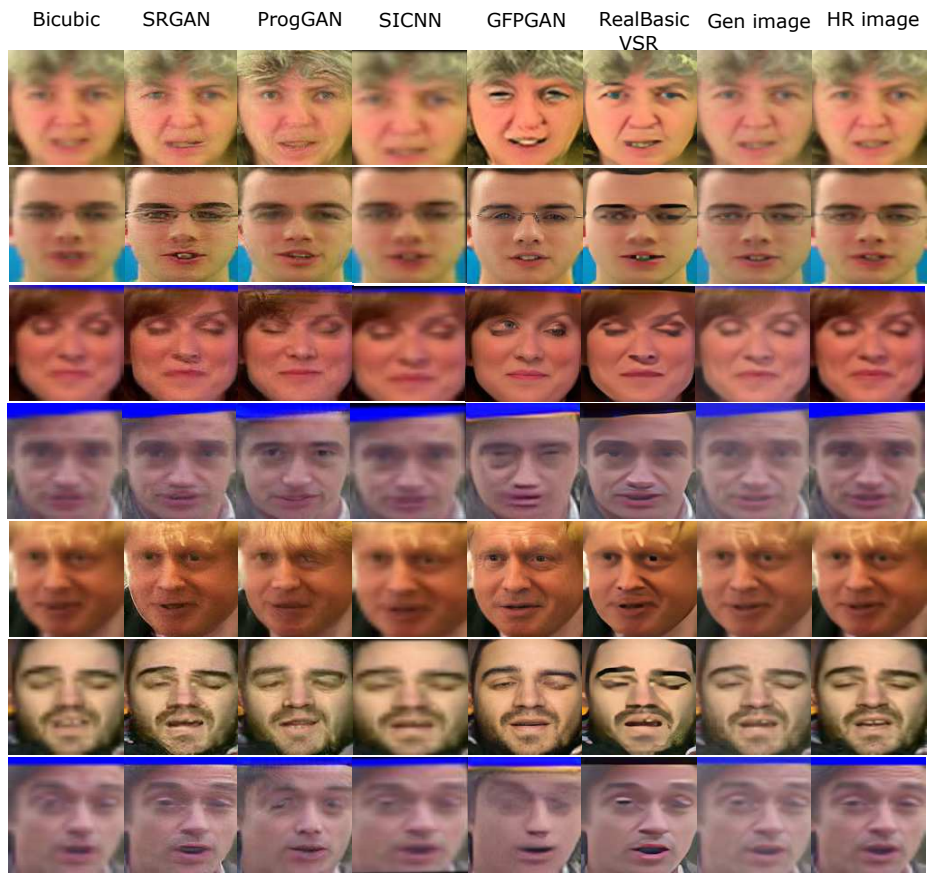}

\caption{Qualitative results comparison with $\times4$ upscaling factor using
various datasets: GRID datset (rows -1 \& 2), LRS2 dataset (rows -
3,4 \& 7) and LRW datset (rows - 5 \& 6).\label{fig:Qualitative-results-comparison}}
\end{figure*}

{\footnotesize{}}
\begin{table*}[h]
\centering
{\footnotesize{}}%
\begin{tabular}{c|c|c|c|c|c|c|c|c|c}
\toprule 
{\footnotesize{}Method} & \multicolumn{3}{c|}{{\footnotesize{}LRW}} & \multicolumn{3}{c|}{{\footnotesize{}LRS2}} & \multicolumn{3}{c}{{\footnotesize{}GRID}}\tabularnewline 
\midrule
 & {\footnotesize{}PSNR} & {\footnotesize{}SSIM} & {\footnotesize{}ERQA} & {\footnotesize{}PSNR} & {\footnotesize{}SSIM} & {\footnotesize{}ERQA} & {\footnotesize{}PSNR} & {\footnotesize{}SSIM} & {\footnotesize{}ERQA}\tabularnewline
\midrule  
{\footnotesize{}Bicubic} & {\footnotesize{}27.734} & {\footnotesize{}0.795} & {\footnotesize{}0.401} & {\footnotesize{}27.991} & {\footnotesize{}0.801} & {\footnotesize{}0.368} & {\footnotesize{}28.192} & {\footnotesize{}0.792} & {\footnotesize{}0.394}\tabularnewline
 
{\footnotesize{}SRGAN} & {\footnotesize{}29.262} & {\footnotesize{}0.865} & {\footnotesize{}0.515} & {\footnotesize{}29.631} & {\footnotesize{}0.842} & {\footnotesize{}0.469} & {\footnotesize{}28.715} & {\footnotesize{}0.848} & {\footnotesize{}0.455}\tabularnewline
 
{\footnotesize{}ProgGAN} & {\footnotesize{}$\mathbf{29.301}$} & {\footnotesize{}0.847} & {\footnotesize{}0.496} & {\footnotesize{}$\mathbf{30.610}$} & {\footnotesize{}0.831} & {\footnotesize{}0.466} & {\footnotesize{}29.988} & {\footnotesize{}0.820} & {\footnotesize{}0.419}\tabularnewline
 
{\footnotesize{}SICNN} & {\footnotesize{}28.031} & {\footnotesize{}0.812} & {\footnotesize{}0.421} & {\footnotesize{}28.172} & {\footnotesize{}0.822} & {\footnotesize{}0.449} & {\footnotesize{}28.042} & {\footnotesize{}0.805} & {\footnotesize{}0.423}\tabularnewline
 
{\footnotesize{}GFPGAN} & {\footnotesize{}28.155} & {\footnotesize{}0.870} & {\footnotesize{}0.490} & {\footnotesize{}29.414} & {\footnotesize{}0.843} & {\footnotesize{}0.485} & {\footnotesize{}28.133} & {\footnotesize{}0.825} & {\footnotesize{}0.424}\tabularnewline
 
{\footnotesize{}RealBasicVSR} & {\footnotesize{}29.370} & {\footnotesize{}0.913} & {\footnotesize{}0.583} & {\footnotesize{}28.595} & {\footnotesize{}0.878} & {\footnotesize{}0.484} & {\footnotesize{}29.930} & {\footnotesize{}0.905} & {\footnotesize{}0.504}\tabularnewline
 
{\footnotesize{}Our method} & {\footnotesize{}28.960} & {\footnotesize{}$\mathbf{0.915}$} & {\footnotesize{}$\mathbf{0.601}$} & {\footnotesize{}28.603} & {\footnotesize{}$\mathbf{0.911}$} & {\footnotesize{}$\mathbf{0.497}$} & {\footnotesize{}$\mathbf{30.791}$} & {\footnotesize{}$\mathbf{0.917}$} & {\footnotesize{}$\mathbf{0.544}$}\tabularnewline
\bottomrule 
\end{tabular}{\footnotesize\par}
{\footnotesize{}\caption{Average PSNR (dB), average SSIM and average ERQA numbers comparison
on various audio-visual datasets.\label{tab:Average-PSNR-(dB),} }
}{\footnotesize\par}
\vspace{-3mm}
\end{table*}
{\footnotesize\par}

For the performance evaluation, we compared our proposed VFHN with
SOTA super resolution methods: SRGAN \cite{ledig2017photo}, ProgGAN \cite{kim2019progressive}, SICNN  \cite{zhang2018super},
GFPGAN \cite{wang2021towards}, RealBasicVSR \cite{chan2021basicvsr} and bicubic interpolation. 

$\mathbf{Grid}$ $\mathbf{speech}$ $\mathbf{corpus}$ $\mathbf{dataset}$:
We evaluated our networks performance with other SOTA
methods using grid speech corpus dataset to obtain quantitative and
visual results. We randomly selected $15$ videos from the dataset
and calculated average SSIM, PSNR and ERQA. Numbers represented in
table \ref{tab:Average-PSNR-(dB),} shows superiority of our network,
since our network achieves highest numbers for all the quantitative
metrics. 

Visual results for the same are shown in Figure \ref{fig:Qualitative-results-comparison}
(rows- 1 \& 2). Facial images generated by SRGAN and ProgGAN have
artifacts present near the teeth, eyes and hair. Whereas SICNN is
producing blurry images. GPFGAN is changing the structure of face
and producing appalling eyes area. Although images generated by RealBasicVSR
looks more sharper than the other methods but they are more looks
like animated pictures rather than real images. Facial images generated
by our method are trying to restore the real information and generating
images similar to the ground-truth image without adding any irrelevant
information. 

$\mathbf{LRS2}$ $\mathbf{dataset}$ and $\mathbf{LRW}$ $\mathbf{dataset}$:
Similarly, we randomly selected 15 videos from both these datsets
and then calculated their average numbers. Our method has achieved
highest average SSIM and ERQA numbers but average PSNR is little less
as compare to other comparison methods. It is proved in previous researches
that we cannot rely on just PSNR in case of super resolution because
high PSNR does not necessarily generates visually more appealing results.
This can also be depicted from Figure \ref{fig:Qualitative-results-comparison}.
Although ProgGAN has the highest average PSNR in both LRS2 and LRW
datasets, yet the generated results by this model have distorted images
(refer Figure \ref{fig:Qualitative-results-comparison} (row 3)),
have various artifacts (refer Figure \ref{fig:Qualitative-results-comparison}
(row 5, 6 \& 7)) and change is shape of mouth region from the original
image (refer Figure \ref{fig:Qualitative-results-comparison} (row
4)). Images generated by using GFPGAN are more inclined toward an
image generation rather than the super resolution task. For example,
the model is adding beard to the face which is not present originally
(refer Figure \ref{fig:Qualitative-results-comparison} (row 6)) and
tries to open the eyes while they are closed in the ground-truth image
(refer Figure \ref{fig:Qualitative-results-comparison} (row 3)).
RealBasicVSR is generating results with unnecessarily sharpening some
parts of face like brows, eyes, nose and wrinkles etc (refer Figure
\ref{fig:Qualitative-results-comparison} (row 4, 5, 6 \& 7)) leading
to generate images looking fake or animated. The images generated
by our model resembles to the ground truth image without any change
in shape and generating realistic texture. 
% \begin{figure*}[ht]
% \centering

% \includegraphics[scale=0.5]{arch_m.eps}
% \caption{Generator architecture. \label{fig:Generator-architecture}}
% \end{figure*}

\section{Conclusion}

In this paper, firstly we explored the semantic relation between audio
waves and corresponding visual frames to maintain temporal consistency
across the frames of videos. We proposed a novel lip-reading loss
inspired by automatic speech recognition. This loss function supervises
the proposed architecture to generate facial images with very fine
texture around the mouth region. To effectively add the salient frequency
features, we added a frequency based loss function in conjunction
with spatial losses. Experimental results show the potential of proposed
model over SOTA methods. 

\bibliographystyle{plain}
\bibliography{ref}
%%%%%%%%% REFERENCES

\end{document}